\documentclass[runningheads]{llncs}

\usepackage[T1]{fontenc}
\usepackage{graphicx}
\usepackage{orcidlink}
\usepackage{multicol}
\usepackage[section]{placeins}
\begin{document}
\title{A Comparison of Fusion Techniques for Multi-Modal Human Activity Recognition on the HARMES Dataset}
\titlerunning{Comparison of Fusion Techniques for Multi-Modal HAR}
%
%
\author{
Ahmed Mohamady\textsuperscript{*}\orcidlink{0009-0006-4721-2731} \and
Robin Burchard\textsuperscript{*}\orcidlink{0000-0002-4130-5287} \and
Kristof Van Laerhoven\orcidlink{0000-0001-5296-5347}
}

\institute{University of Siegen, Germany \\
\email{engahmedmohamady23@gmail.com}\\
\url{https://ubicomp.eti.uni-siegen.de/}
\\[1ex]
\textsuperscript{*}These authors contributed equally to this work.
}
\authorrunning{Mohamady and Burchard et al.}
%

%
\maketitle              
\begin{abstract}
Recent advances in Human Activity Recognition (HAR) from wearable sensors have shown that multi-modal deep learning models consistently outperform their uni-modal counterparts. Modalities can include IMUs, RGB cameras, audio signals, and others. One important aspect of multi-modal deep learning is the sensor fusion approach we apply. Over recent years, multiple fusion paradigms have been proposed for multi-modal HAR. However, to the best of our knowledge, no head-to-head comparison of these paradigms exists on a common multi-modal HAR benchmark dataset. To address this research gap, we systematically compare seven state-of-the-art sensor fusion methods on the recently released HARMES dataset, which comprises 61 hours of fully labeled IMU, audio, and ambient humidity data. The chosen dataset focuses on 15 household and personal hygiene activities of daily living (ADLs). By applying the seven different fusion techniques to a state-of-the-art multi-modal model architecture, we show that Gated Multi-modal Fusion achieves the highest macro F1-score (0.82), surpassing the concatenation-based late fusion HARMES paper baseline of 0.76 by +6pp under leave-one-participant-out evaluation. All code used in our experiments is made publicly available on GitHub\footnote{https://github.com/AhmedMohamady98/A-Comparison-of-Fusion-Techniques-for-Multi-Modal-Human-Activity-Recognition-on-the-HARMES-Dataset}.

\end{abstract}
\section{Introduction}
Human Activity Recognition (HAR) from wearable and ambient sensors is a
foundational task for healthcare monitoring, assistive living, smart-home automation, and activity-aware computing \cite{bullingTutorialHumanActivity2014}. The past decade has seen rapid progress driven by deep learning architectures applied to inertial measurement unit (IMU) data \cite{wangDeepLearningSensorbased2019,ordonezDeepConvolutionalLSTM2016}.
These models now routinely achieve high classification accuracy on scripted
activity corpora, but they operate on a single sensory channel and remain fragile to real-world confounds such as handedness, sensor placement variability, and activities with similar motion signatures.\\
A growing body of work shows that the limitations of unimodal HAR can be overcome by combining multiple sensing modalities \cite{ramachandramDeepMultimodalLearning2017,atreyMultimodalFusionMultimedia2010}. For example, audio captures characteristic acoustic signatures of water-based and mechanical activities (e.g., dishwashing, vacuuming, brushing teeth) that are difficult to disambiguate from IMU sensors alone. Environmental sensors such as humidity provide contextual priors: high humidity near a sink is indicative of water-related tasks \cite{burchardMultimodalAtmosphericSensing2025,burchardImprovedStrategiesMultimodal2026}. The HARMES dataset \cite{burchardHARMESMultiModalDataset2026a} offers a public, multi-participant, long-duration benchmark with synchronised IMU, audio, and humidity streams recorded from activities of daily living (ADLs), enabling multi-modal fusion research on ecologically valid data. The literature on multi-modal fusion is diverse. Architectures range from simple late concatenation of modality embeddings to gated units that learn per-sample modality weights \cite{arevaloGatedMultimodalNetworks2020}, bottleneck transformers routing information through shared latent tokens \cite{nagraniAttentionBottlenecksMultimodal2022}, directional cross-attention \cite{tsaiMultimodalTransformerUnaligned2019},
low-rank tensor factorization \cite{liuEfficientLowrankMultimodal2018}, and decision-level
weighted averaging. While these approaches beat the respective baselines, each approach has been validated on different datasets with different encoder backbones and evaluation protocols, making direct comparison across many fusion methods impossible. A researcher choosing a fusion strategy for a new multi-modal HAR system currently has no direct empirical guidance. In this work, we provide such a comparison. Using HARMES as a shared testbed, we train and evaluate \textbf{seven fusion strategies} under identical conditions: the same three encoder backbones, the same 10-second window size, the same training hyperparameters, and the same evaluation metrics. We report results under 3-fold group cross-validation and 20-fold LOPO evaluation, matching the HARMES benchmark protocol. We further conduct a modality ablation and a subgroup analysis of left-handed participants.

Our contribution is threefold. 
\begin{enumerate}
    \item We present a systematic head-to-head comparison of seven prominent multi-modal fusion paradigms on a
uniform HAR benchmark, providing an empirical ranking independent of backbone or protocol variation.
    \item We identify the practical drivers of fusion performance on the HARMES dataset by identifying the per-class-performance contribution of each modality.
    \item We show that multi-modal fusion partially mitigates the handedness gap by relying on handedness-invariant audio cues when IMU signals become unreliable.
\end{enumerate}

\section{Related Work}
To position this work in the context of current research, we discuss existing HAR model architectures, fusion methods, and fusion method comparison studies.

\subsection{Encoder Architectures for Multi-Modal Wearable HAR}

Deep learning for IMU-based activity recognition has progressed through several generations of architecture. Ord\'o\~nez et al.~\cite{ordonezDeepConvolutionalLSTM2016} introduced DeepConvLSTM, pairing convolutional feature extraction with LSTM temporal modelling into the baseline most later work is measured against, while Ha and Choi \cite{haConvolutionalNeuralNetworks2016} apply CNNs that can generalise across sensor placements via partial weight sharing. The trend has since
shifted toward lightweight, deployable designs. Zhou et al.~\cite{zhouTinyHARLightweightDeep2022} proposed TinyHAR, which chains
depthwise separable convolutions with transformer encoded blocks, LSTM, and temporal attention, while remaining small enough for execution on mobile devices. Bian et al.~\cite{bianTinierHARUltraLightweightDeep2025} published an even smaller, transformer-free model with TinierHAR, which is built around bi-directional GRUs and temporal aggregation instead, reducing parameter count and computational cost significantly while maintaining competitive accuracy.

Audio is a powerful complementary channel for a multitude of activities. Gong
et al.~\cite{gongASTAudioSpectrogram2021} adapted the Vision Transformer to audio by treating
log-mel spectrograms as patch sequences, producing the Audio Spectrogram
Transformer (AST), which remains a strong and widely used audio classification backbone. Mollyn et al.~\cite{mollynSAMoSASensingActivities2022} showed that IMU-triggered, privacy-preserving audio sampling recognizes 26 daily activities without continuous recording, and Lee et al.~\cite{leeMultimodalSensorFusion2024} confirmed that audio is highly effective for detecting activities with minimal limb motion. For general timeseries signals, Ekambaram et al.~\cite{ekambaramTSMixerLightweightMLPMixer2023} proposed TSMixer, an all-MLP time-series architecture.

Several multi-modal HAR systems combine these channels in different ways: E.g., AttnSense's multi-level attention \cite{maAttnSenseMultilevelAttention2019}, Cosmo's quality-guided contrastive fusion \cite{ouyangCosmoContrastiveFusion2022}, and Mamba-MHAR's two-branch selective state space models \cite{leMambaMHAREfficientMultimodal2025}.

\subsection{Multi-Modal Fusion Strategies}
\label{sec:related_architectures}
Multi-modal fusion can be placed along a spectrum defined by how early in the
processing pipeline modalities are combined \cite{atreyMultimodalFusionMultimedia2010,ramachandramDeepMultimodalLearning2017}.
At one extreme, early fusion merges raw or low-level features before any
modality-specific learning has taken place. At the other extreme, decision-level
fusion combines the outputs of fully independent per-modality classifiers. Between
these two there exists a multitude of intermediate approaches that operate on
learned modality embeddings. Ramachandram and Taylor \cite{ramachandramDeepMultimodalLearning2017} identified
at least five distinct architectural families within this space, and the field has
continued to diversify since. Below, we list and discuss the most prominent fusion approaches in current literature.

\textbf{Early fusion} concatenates raw sensor readings into a single input vector
before any model processing. It is simple but assumes all modalities share the same
temporal resolution and scale, which rarely holds for heterogeneous sensors such as
IMUs, microphones, and environmental probes. Yilmaz. et al.~\cite{yilmazHierarchicalHumanActivity2025} show that audio and IMU data can be processed into images, which leads to beneficial results with their early fusion approach. In practice, early fusion is often outperformed by
methods that encode each modality separately \cite{ramachandramDeepMultimodalLearning2017,atreyMultimodalFusionMultimedia2010,munznerCNNbasedSensorFusion2017}. 

\textbf{Late embedding fusion}, oftentimes called feature-level or representation-level fusion, concatenates the output embeddings of per-modality encoders and feeds the combined vector to a shared classifier. We refer to embedding-level feature fusion as late embedding fusion to distinguish it from decision-level late fusion. It is often applied in 
multi-modal HAR because it decouples encoder training from fusion: each encoder can
be pre-trained independently, and the fusion head stays lightweight. M\"unzner
et al.~\cite{munznerCNNbasedSensorFusion2017} found embedding concatenation outperformed
decision-level fusion on PAMAP2, and it has since been used for HAR tasks, e.g., household activity
recognition \cite{burchardHARMESMultiModalDataset2026a} and exercise
counting \cite{leeMultimodalSensorFusion2024}. 

\textbf{Gated fusion} addresses the fact that not all modalities are equally
informative for every sample: a static concatenation head cannot down-weight a
noisy input. Arevalo et al.~\cite{arevaloGatedMultimodalNetworks2020} introduced the Gated
Multi-modal Unit (GMU), where a sigmoid gate computed from all modalities controls
how much each contributes to the fused vector. Because the gate is conditioned on
the combined context, the model can learn to suppress modalities adaptively, when they are less informative, e.g., due to background noise. As the weights for each modality can be inspected, GMU produces interpretable modality ``importances''.

\textbf{Tensor fusion} models cross-modal interactions explicitly via the outer
product of modality embeddings, capturing pairwise and higher-order combinations in
one operation. The Tensor Fusion Network (TFN) of Zadeh
et al.~\cite{zadehTensorFusionNetwork2017} showed this improves over additive baselines on
sentiment analysis, but the outer-product tensor grows multiplicatively with the
number and dimensionality of modalities, making it expensive. Liu
et al.~\cite{liuEfficientLowrankMultimodal2018} addressed this with Low-Rank multi-modal Fusion
(LMF), which decomposes the weight tensor into modality-specific low-rank factors
combined by an element-wise product. LMF matches or beats TFN with far fewer
parameters, showing the full-rank tensor is largely redundant.

\textbf{Cross-modal attention} builds on the Transformer \cite{vaswaniAttentionAllYou2017}, using one modality as the query and another as the value so the query can attend to the most relevant parts of the other modality. Tsai et al.~\cite{tsaiMultimodalTransformerUnaligned2019} systematised this as the multi-modal Transformer (MulT), applying it in both directions for every modality pair, which gives six cross-attention streams per layer with three modalities. MulT is commonly used for language-vision-acoustic fusion. The same idea shows up in ViLBERT's co-attentional layers \cite{luViLBERTPretrainingTaskAgnostic2019} and in RGB-D detection \cite{wangCMASODCrossmodalAttention2025}. Within HAR, related attention-based fusion ideas appear in AttnSense \cite{maAttnSenseMultilevelAttention2019}, which applies multi-level attention over inertial sensors, implicitly weighting both on-body placement channels and time steps

\textbf{Bottleneck transformer fusion} is a compute-efficient alternative to full
cross-attention. Instead of letting every token attend to every other, Nagrani
et al.~\cite{nagraniAttentionBottlenecksMultimodal2022} route all cross-modal information through a
small set of shared bottleneck tokens. In the multi-modal Bottleneck Transformer
(MBT), each layer processes a modality's tokens together with the bottleneck, but
never mixes two modalities directly. The per-modality bottleneck copies are then
averaged into a shared state. On AudioSet and VGGSound, it beat late fusion and full
cross-attention with fewer parameters, and the bottleneck may also act as an implicit regularizer by constraining cross-modal information flow to a limited set of shared tokens. Bralina et al.~\cite{bralinaAdaptiveBottleneckTransformer2026} extended it with the Adaptive Bottleneck Transformer (AMBT), adding factorised contrastive learning to prevent modality collapse.

\textbf{Decision-level fusion} never builds a shared representation. Each modality has its own classifier producing a class distribution, and these get combined by a fixed rule or a learned weighted sum. Atrey et al.~\cite{atreyMultimodalFusionMultimedia2010} formalised this as one of the three canonical fusion levels. The approach is modular and transparent, since the weights directly quantify per-modality confidence, but it loses interactions at the representation level: the model has no way to discover that a humid environment combined with repetitive arm motion is more diagnostic than either signal on its own.

\textbf{CLS-token fusion} treats each per-modality embedding as a token in a shared
sequence with a prepended, learnable \texttt{[CLS]} token, processed by standard
transformer encoder layers \cite{vaswaniAttentionAllYou2017} following the
classification convention of BERT (Devlin et al.~\cite{devlinBERTPretrainingDeep2019}) and ViT (Dosovitskiy et al.~\cite{dosovitskiyImageWorth16x162021}). The
\texttt{[CLS]} state aggregates all modality tokens through unrestricted
self-attention. There is no bottleneck constraint as in MBT, and no directional
structure as in MulT. All tokens take part in one unified attention operation.

\textbf{Contrastive self-supervised fusion} is a term for self-supervised fusion methods that align modalities using the natural
pairing structure of the data rather than merging them into one representation. Tian
et al.~\cite{tianContrastiveMultiviewCoding2020} introduced Contrastive Multiview Coding (CMC),
pulling together views of the same scene and pushing apart different scenes, and
Radford et al.~\cite{radfordLearningTransferableVisual2021} scaled this to image-text pairs with
CLIP, enabling zero-shot transfer via similarity retrieval. Girdhar
et al.~\cite{girdharImageBindOneEmbedding2023} extended it to six modalities, IMU included,
with ImageBind, showing that pairing each modality with images alone suffices to
align them all. Closest to our setting, Moon et al.~\cite{moonIMU2CLIPLanguagegroundedMotion2023}
introduced IMU2CLIP, aligning wrist-worn IMU with video and text, and Ouyang
et al.~\cite{ouyangCosmoContrastiveFusion2022} applied contrastive fusion to wearable HAR with
Cosmo. These approaches suit wearable sensing well, where labels are expensive but
paired unlabeled recordings are cheap to collect, but do not apply to our fully supervised modality fusion problem.

\textbf{Feature conditioning} uses one modality to modulate the internal computation
of a network processing another, instead of merging them at a fixed layer. Perez
et al.~\cite{perezFiLMVisualReasoning2018} introduced Feature-wise Linear Modulation (FiLM), where
a conditioning network predicts per-channel scale and shift parameters applied to a
target network's feature maps. It was proposed for visual question answering and has
since been used wherever one signal provides context for another. Rahman
et al.~\cite{rahmanIntegratingMultimodalInformation2020} adapted this to pretrained transformers with
the multi-modal Adaptation Gate (MAG), injecting a learned shift into BERT's
representations from visual and acoustic signals while leaving the backbone intact.

\textbf{Latent bottleneck architectures} generalise the MBT bottleneck into a
modality-agnostic framework. Jaegle et al.~\cite{jaeglePerceiverGeneralPerception2021} introduced
the Perceiver, which maps inputs of arbitrary modality and size into a fixed-size
latent array via asymmetric cross-attention, then processes that array with standard
Transformer layers. Decoupling compute cost from input size lets one architecture
handle images, audio, point clouds, and video with minimal modality-specific architectural changes. Where MBT still feeds per-modality token sequences in separately, the Perceiver compresses the entire input into a shared latent space from the first step.

\textbf{Feature recalibration} lets each modality adjust the internal features of
the others while the per-modality streams stay separate. Joze
et al.~\cite{vaezijozeMMTMMultimodalTransfer2020} introduced the multi-modal Transfer Module (MMTM), which
uses squeeze-and-excitation to recalibrate the channel-wise features of each
convolutional stream from a joint summary of all modalities. Because it slots
between existing branches, each one can keep its pretrained weights. Within HAR, Gao
et al.~\cite{gaoMMTSAMultiModalTemporal2023} proposed MMTSA, which encodes inertial signals as
Gramian Angular Field images and fuses RGB and IMU streams via inter-segment
attention, reaching strong cross-participant accuracy.

\subsection{Fusion Comparison Studies}

The literature on multi-modal fusion is large, but systematic head-to-head comparisons of multiple fusion strategies under controlled conditions are surprisingly rare. Most papers introduce a single new method and compare it against a concatenation baseline, making it difficult to draw conclusions about the relative merit of different architectural families.

At the survey level, Atrey et al.~\cite{atreyMultimodalFusionMultimedia2010} provide a
foundational taxonomy of fusion approaches, distinguishing feature-level,
decision-level, and hybrid fusion and categorising methods by their combination
rule. Ramachandram and Taylor \cite{ramachandramDeepMultimodalLearning2017} later surveyed the
deep multi-modal learning landscape and identified five architectural families, but
their treatment is conceptual rather than empirical. In the HAR domain
specifically, Qiu et al.~\cite{qiuMultisensorInformationFusion2022} reviewed over 300
multi-sensor papers and mapped out dominant design patterns, and Yadav
et al.~\cite{yadavReviewMultimodalHuman2021} concluded that multi-modal fusion consistently
outperforms unimodal approaches. Neither study, however, evaluates two or more
fusion methods under the same experimental conditions.

One of the few HAR papers to directly compare fusion strategies is
M\"unzner et al.~\cite{munznerCNNbasedSensorFusion2017}, who tested sensor-level CNN
fusion against decision-level fusion on the PAMAP2 dataset and reported higher performance for feature-level fusion than decision-level fusion. Aguileta et al.~\cite{aguiletaMultiSensorFusionActivity2019} similarly
surveyed multi-sensor fusion for activity recognition and noted that no consensus
had emerged on the best strategy, with results varying substantially across
datasets and sensor combinations.

Outside HAR, more controlled comparisons do exist but remain tied to their own domains and encoder choices. Tsai et al.~\cite{tsaiMultimodalTransformerUnaligned2019} evaluated MulT against late fusion and tensor fusion on language-vision-acoustic sentiment datasets. Nagrani et al.~\cite{nagraniAttentionBottlenecksMultimodal2022} compared MBT against late fusion and full cross-attention on video-audio classification, finding that the bottleneck constraint does not hurt performance while reducing computational cost. Koutoupis et al.~\cite{koutoupisMoreMerrierContrastive2026} evaluated contrastive fusion objectives for higher-order multi-modal alignment across several non-HAR settings. These studies are valuable but their findings do not transfer directly to wearable sensor streams, where modality characteristics, sampling rates, and temporal structure differ fundamentally from video, language, and audio-visual pairs.

To the best of our knowledge, we found no prior study that compares such a broad range of fusion paradigms under identical encoder backbones on a public multi-modal HAR benchmark.

\section{Methodology}
\label{sec:Methdology}

\subsection{Dataset}
We use the HARMES dataset \cite{burchardHARMESMultiModalDataset2026,burchardHARMESMultiModalDataset2026a}, a public multi-modal corpus
for wearable HAR. It contains recordings from 20 participants (10 female, 10 male,
ages 18--67, mean~37.8, SD~14.4) performing 15 activities of daily living in their own
homes. Each participant contributed about three hours of fully labeled data across
three recording sessions. We exclude the fourth, free-form, session, leaving 61 hours
of labelled data in total.

\subsubsection{Modalities}
Three synchronized streams are provided. The \textit{IMU} stream includes dual-wrist
accelerometer and gyroscope data at 50~Hz, from a Puck.js device on the left wrist
and a smartwatch on the right, totaling 12 channels (3-axis acceleration and 3-axis
angular velocity per wrist). The \textit{audio} stream is an ambient recording at
44.1~kHz from a wrist-worn microphone (in the smartwatch), with no speech contained in the dataset for privacy reasons. The
\textit{atmospheric} stream is humidity from a wrist-mounted BME280 sensor at 1~Hz,
up-sampled to the IMU grid as a 1D time series.

\subsubsection{Activity classes}
The 15 labelled activities fall into three groups: self-care (washing hands,
brushing teeth, applying hand cream, disinfecting hands), household (floor cleaning,
window cleaning, vacuum cleaning, washing dishes, putting away dishes, cleaning
table, cleaning out dishwasher), and feeding (making tea, cutting vegetables,
drinking, watering plants). The Null class covers background and transition
periods between activities, leading to a total of 16 classes. At 10~s granularity,
the Null class is the most common label (20.5\% of all windows).

\subsubsection{Left-handed participants}
Three participants (P07, P10, P14) are left-handed, a 15\% prevalence in line with
the real-world estimate of 9--18\% \cite{burchardHARMESMultiModalDataset2026a}. This leads to a
generalization challenge: IMU signals depend on which hand performs the dominant
motion. While better IMU sensor placement generalization can likely be achieved with data augmentation, audio is expected to be mostly handedness-invariant.

\subsection{Encoder Architectures}
We process each modality with its own encoder that maps the raw stream to a
128-dimensional embedding. We fix this embedding size across all three encoders so
that every fusion method operates on a common representation, which lets us swap
fusion strategies without changing the encoders. The IMU and humidity encoders are
trained from scratch, while the audio encoder uses a frozen pretrained backbone with
only a small trainable head.
\subsubsection{IMU encoder}
For the dual-wrist inertial stream, we use TinyHAR \cite{zhouTinyHARLightweightDeep2022}. It first
applies lightweight convolutions across the sensor channels to extract local motion
patterns, then a temporal self-attention block to capture how those patterns relate
over the length of a window. The classic baseline here is
DeepConvLSTM \cite{ordonezDeepConvolutionalLSTM2016}, which pairs convolutions with a recurrent
layer, but its LSTM processes time steps sequentially and is comparatively heavy to
train. TinyHAR reaches similar or better accuracy on standard HAR benchmarks at a
fraction of the parameters, and its attention block handles long-range temporal
structure without the sequential bottleneck of a recurrent network. Recent
state-space models such as HARMamba \cite{liHARMambaEfficientLightweight2025} are promising but less
established, so we favour TinyHAR as a proven, compact backbone well suited to
wrist-worn data.
\subsubsection{Audio encoder}
For the ambient audio stream, we use the Audio Spectrogram Transformer
(AST) \cite{gongASTAudioSpectrogram2021}, which treats a spectrogram as a sequence of patches and
applies transformer attention over them. We keep its backbone frozen and train only
a small projection head. The reason is transfer learning: AST is pretrained on
AudioSet, a very large and diverse sound corpus, so it already carries a strong
general audio representation. Our labelled HAR audio is far too small to train a
comparable model from scratch, and fine-tuning the whole backbone on it would risk
overfitting. Convolutional audio models such as VGGish or PANNs were possible
alternatives, but AST's attention gives a stronger pretrained starting point and a
clean way to freeze the backbone and adapt with a single linear layer.
\subsubsection{Humidity encoder}
For the atmospheric stream, we use TSMixer \cite{ekambaramTSMixerLightweightMLPMixer2023}, an all-MLP architecture that alternates between mixing information along the time axis and along the feature axis. Humidity is a slow, single-channel signal, so it does not need the heavy machinery of attention or recurrence to be modelled well. A transformer would be overkill and prone to overfitting on such a simple stream, while an LSTM adds sequential cost for little benefit. TSMixer stays lightweight and trains quickly, which keeps the humidity branch cheap relative to the IMU and audio encoders, an appropriate balance given that humidity likely carries the least activity-discriminative information of the three \cite{burchardHARMESMultiModalDataset2026a}.

\subsection{Fusion Architectures}
\label{sec:met_fusion_arch}

We compare seven fusion methods drawn from the strategies surveyed in
Section \ref{sec:related_architectures}. Our aim is not to propose a new fusion method but to
compare existing ones under identical conditions, so the selection follows a single
principle: every method must operate on the fixed 128-dimensional embeddings
produced by our encoders, treat the three modalities symmetrically, and require no
modality-specific pretraining or input format. This allows us to use the same encoders for every fusion method and vary only the fusion step, so any difference in performance is attributable to the fusion mechanism alone.

This criterion rules out several families discussed in Section \ref{sec:related_architectures}.
Contrastive and joint-embedding methods such as CMC, CLIP, ImageBind, and IMU2CLIP
need large-scale paired data and a separate pretraining stage, which does not fit a
controlled supervised comparison on a single dataset. Methods tied to a particular
input structure, such as the Perceiver's raw-input bottleneck or MMTM and MMTSA's
convolutional streams, would bind the comparison to a specific encoder format. Conditioning methods such as FiLM and MAG assume an asymmetric relationship in which one modality modulates another, which does not match three
co-equal sensor streams.

The seven retained methods cover the remaining intermediate and decision-level
paradigms that do fit this setting (cf. Section \ref{sec:related_architectures} for the exact descriptions):

\begin{itemize}
  \item \textbf{Late Fusion}: embedding concatenation, followed by a simple MLP classification head.
  \item \textbf{Gated Multi-modal Fusion (GMF)}: learned per-modality
        gating.
  \item \textbf{Low-Rank multi-modal Fusion (LMF)}: tensor interaction
        with a low-rank factorisation.
  \item \textbf{Cross-Modal Attention (CMA)}: directional attention
        between every modality pair.
  \item \textbf{Multi-modal Bottleneck Transformer (MBT)}: cross-modal
        exchange through shared bottleneck tokens.
  \item \textbf{CLS-Token Transformer}: a learnable token aggregating all
        modalities under unrestricted self-attention.
  \item \textbf{Decision Fusion}: a learned weighted sum of per-modality
        class distribution predictions.
\end{itemize}

\begin{figure}
    \centering
    \includegraphics[width=1\linewidth]{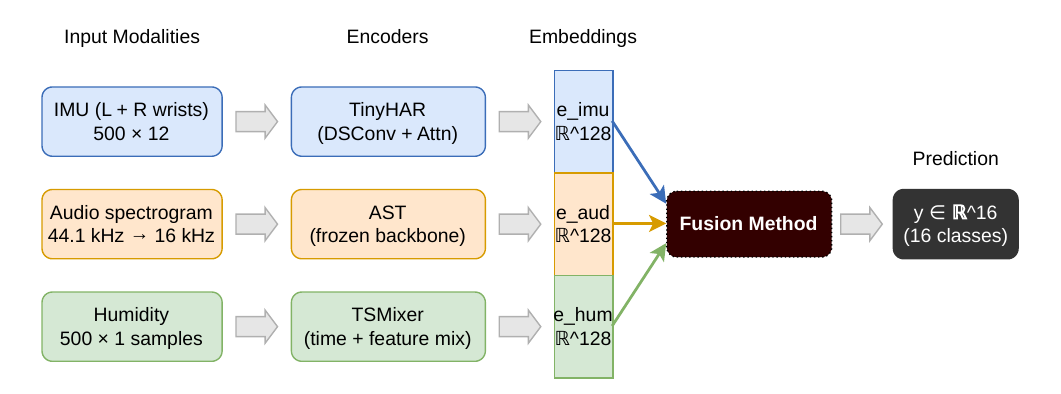}
    \caption{Overview of the multi-modal pipeline. Each modality is encoded into a
  128-dimensional embedding by a dedicated encoder. The three embeddings are then
  combined by one of seven interchangeable fusion methods to predict the activity
  class. The fusion block is held generic here. The seven concrete architectures are
  detailed in Section~\ref{sec:met_fusion_arch} and Appendix \ref{asec:fusion_strategies}.}
    \label{fig:fusion_architecture_overview}
\end{figure}

Together, these span the breadth of the fusion taxonomy, from the simplest
concatenation to tensor, gated, attention-based, and bottleneck designs. Six operate
at the intermediate (embedding) level and one, Decision Fusion, at the decision
level, so the set also covers two of the three canonical fusion stages. Early
(raw-signal) fusion is excluded by construction, since all methods operate on encoder
embeddings. The general structure of our multi-modal classification pipeline is displayed in Fig. \ref{fig:fusion_architecture_overview}.
Full architectural details for each method are given in the respective original publications and in graph-based visual representations in Appendix \ref{asec:fusion_strategies}.

\subsection{Training and Evaluation Protocol}

\subsubsection{Windowing}
We segment the continuous recordings into 10-second windows (leading to 500 samples per window at 50~Hz).
For 3-fold cross-validation, we use a step of 250 samples (50\% overlap), which gives
roughly 43{,}760 windows. For LOPO we, use non-overlapping windows (step~=~500,
matching the HARMES paper \cite{burchardHARMESMultiModalDataset2026a}), yielding 21{,}897 windows. Each
window takes the activity label that covers most of its span (simple majority voting). Because the sliding
window runs across the entire recording, including background and transition periods,
windows without majority activity labels are assigned to the ``Null'' class.

\subsubsection{Cross-validation splits}
We use two cross-validation protocols: 3-fold group cross-validation for all fusion methods and 20-fold Leave-One-Participant-Out for the best method from the 3-fold CV evaluation, for comparison with the HARMES baseline. In \textbf{3-fold group cross-validation},
participants are split into three disjoint folds (see Table \ref{tab:folds}). Within each
fold, the smallest-ID participant outside the test set is held out for validation,
used for early stopping and checkpoint selection, and the rest form the training
pool. Train, validation, and test sets are disjoint at the participant level in every
fold. In \textbf{20-fold Leave-One-Participant-Out (LOPO)}, each participant is held
out once as the test set. For each test participant $P_k$, the smallest-ID participant
among the remaining 19 is used for validation, and the other 18 are used for training. The LOPO paradigm
matches the evaluation in the original HARMES paper \cite{burchardHARMESMultiModalDataset2026a},
allowing a direct comparison. We use three folds rather than a larger fold count for the main comparison for experiment runtime reasons. As every fusion method is trained and evaluated on each fold, the total training cost
scales with the number of methods times the number of folds. Three folds still leave
roughly seven held-out participants per fold, enough for a stable estimate, while
keeping the seven-way comparison tractable. We reserve the more expensive LOPO
protocol for the single best-performing method, where a direct comparison with the
original HARMES paper is most informative.

\begin{table}
\centering
\caption{Participant assignment for the three cross-validation folds.}
\label{tab:folds}
\begin{tabular}{cccc}
\hline
Fold & Test & Val & Train \\
\hline
0 & P1--P7  & P8 & P9--P20 \\
1 & P8--P14 & P1 & P2--P7, P15--P20 \\
2 & P15--P20 & P1 & P2--P14 \\
\hline
\end{tabular}
\end{table}

\subsubsection{Training configuration}
We apply an almost identical training configuration to all model configurations. We use Adam as the optimizer with cosine annealing (no restarts), a maximum of 50 epochs, and a batch size of 32. The learning rate is $10^{-3}$ for every method except LMF, which uses
$5\times10^{-3}$ to compensate for vanishing gradients through its three-way tensor
product. Early stopping (patience 10 epochs) monitors validation macro F1, and the best
checkpoint is loaded for a single final evaluation on the held-out test set, which is
never used for model selection. TinyHAR and TSMixer are trained from scratch, while
the AST backbone, which is available pre-trained, stays frozen. We train all methods using cross-entropy loss except Decision
Fusion, for which we use negative log-likelihood on log-probabilities.


\subsubsection{Evaluation}
For our comparison, we compute metric scores on ground truth labels and predictions by each model.
The primary metric is macro-averaged F1, which weights all 16 classes equally
regardless of their frequency. This is relevant, as the class distribution in the original dataset is imbalanced. We also report accuracy for comparability with prior work, along with per-class and per-participant results, and confusion matrices for diagnostic analysis of confounded classes/samples.

\section{Results}
\label{sec:results}

We compare the seven fusion methods against the unimodal baselines and the HARMES
reference using the protocols of Section \ref{sec:Methdology}. Our analysis, and thus, the results are organised
around two questions: how much does combining modalities help, and which fusion
strategy makes the best use of them. Table \ref{tab:master} summarises the aggregated results for each fusion strategy, and the subsections that follow break them down by class, by participant, and by handedness.

Fusion strategies improve over the strongest single modality across the board. The best
unimodal model is AST on audio at 0.734 macro F1, and the best fusion method, GMF,
reaches 0.827, an improvement of 9.3 points. Audio is by far the most informative
single stream, while humidity on its own is close to the chance level at 0.09 macro F1, which
is consistent with the modality ablation reported in the HARMES paper \cite{burchardHARMESMultiModalDataset2026a}. The 3-fold and
leave-one-participant-out protocols agree closely for GMF (0.827 and 0.819), so the comparison is stable across evaluation schemes, and GMF under LOPO also exceeds the HARMES multi-modal baseline of 0.760 by 5.9 points.

\begin{table}[t]
\centering
\caption{Results across modality combination, fusion strategies, and cross-validation modes. F1 is the macro F1 score, averaged over folds (3-fold CV) or participants (LOPO), Acc. is the prediction accuracy, averaged in the same way. All fusion methods use the three modalities jointly (Modality = All). \textbf{Bold} font marks the
best result within each section. The HARMES baseline is the multimodal
leave-one-participant-out macro F1 reported by Burchard et al.~\cite{burchardHARMESMultiModalDataset2026a}}
\label{tab:master}
\begin{tabular}{llcc}
\hline
Model / Method & Modality & F1 & Acc \\
\hline
\multicolumn{4}{l}{\textit{Unimodal baselines (3-fold CV)}} \\
TinyHAR        & IMU      & 0.696          & 0.724 \\
\textbf{AST}   & Audio    & \textbf{0.734} & \textbf{0.777} \\
TSMixer        & Humidity & 0.088          & 0.210 \\
\hline
\multicolumn{4}{l}{\textit{Fusion methods (3-fold CV)}} \\
\textbf{GMF}     & All & \textbf{0.827} & \textbf{0.854} \\
Late Fusion      & All & 0.817          & 0.845 \\
CMA              & All & 0.795          & 0.831 \\
CLS Transformer  & All & 0.793          & 0.832 \\
MBT              & All & 0.787          & 0.821 \\
LMF              & All & 0.747          & 0.786 \\
Decision Fusion  & All & 0.747          & 0.783 \\
\hline
\multicolumn{4}{l}{\textit{Leave-one-participant-out (LOPO)}} \\
\textbf{GMF}                                & All & \textbf{0.819} & \textbf{0.856} \\
HARMES baseline \cite{burchardHARMESMultiModalDataset2026a} & All & 0.760          & 0.794 \\
\hline
\end{tabular}

\end{table}

\subsection{Fusion Method Comparison}
\label{sec:res-fusion}

\begin{figure}[h!]
  \centering
  \includegraphics[width=\linewidth]{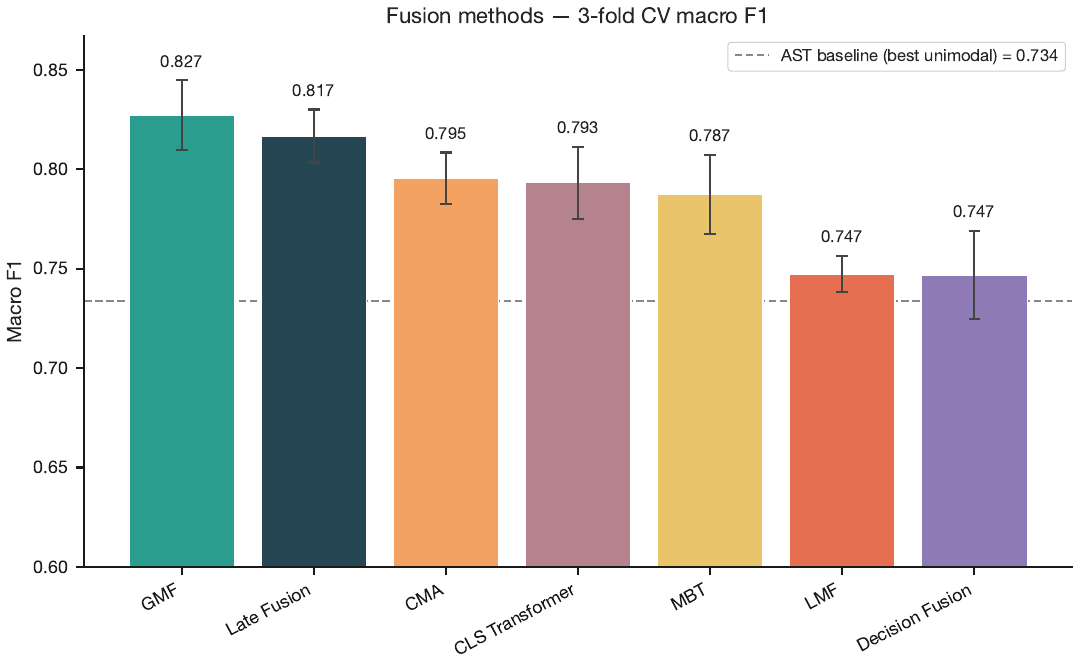}
  \caption{Fusion method comparison on 3-fold group cross-validation (macro F1).
  All methods use the three modalities jointly. The dashed line marks the best
  unimodal baseline (AST).}
  \label{fig:leaderboard}
\end{figure}

Figure \ref{fig:leaderboard} ranks the seven fusion methods on 3-fold cross-validation.
Every method clears the best unimodal baseline (AST, 0.734 macro F1), confirming that
combining modalities helps regardless of the tested fusion strategy. The margin, however,
varies. GMF leads at 0.827, with Late Fusion close behind at 0.817. The two
simplest mechanisms, gated summation and plain concatenation, are also the two
strongest. The attention- and transformer-based methods form a middle cluster (CMA
0.795, CLS-Token Transformer 0.793, MBT 0.788) that lands several points below GMF.
LMF and Decision Fusion trail at 0.747, beating the audio-based unimodal baseline only by a smaller margin.

\subsection{Class performance analysis}
\label{sec:res-perclass}

\begin{figure}[h!]
  \centering
  \includegraphics[width=\linewidth]{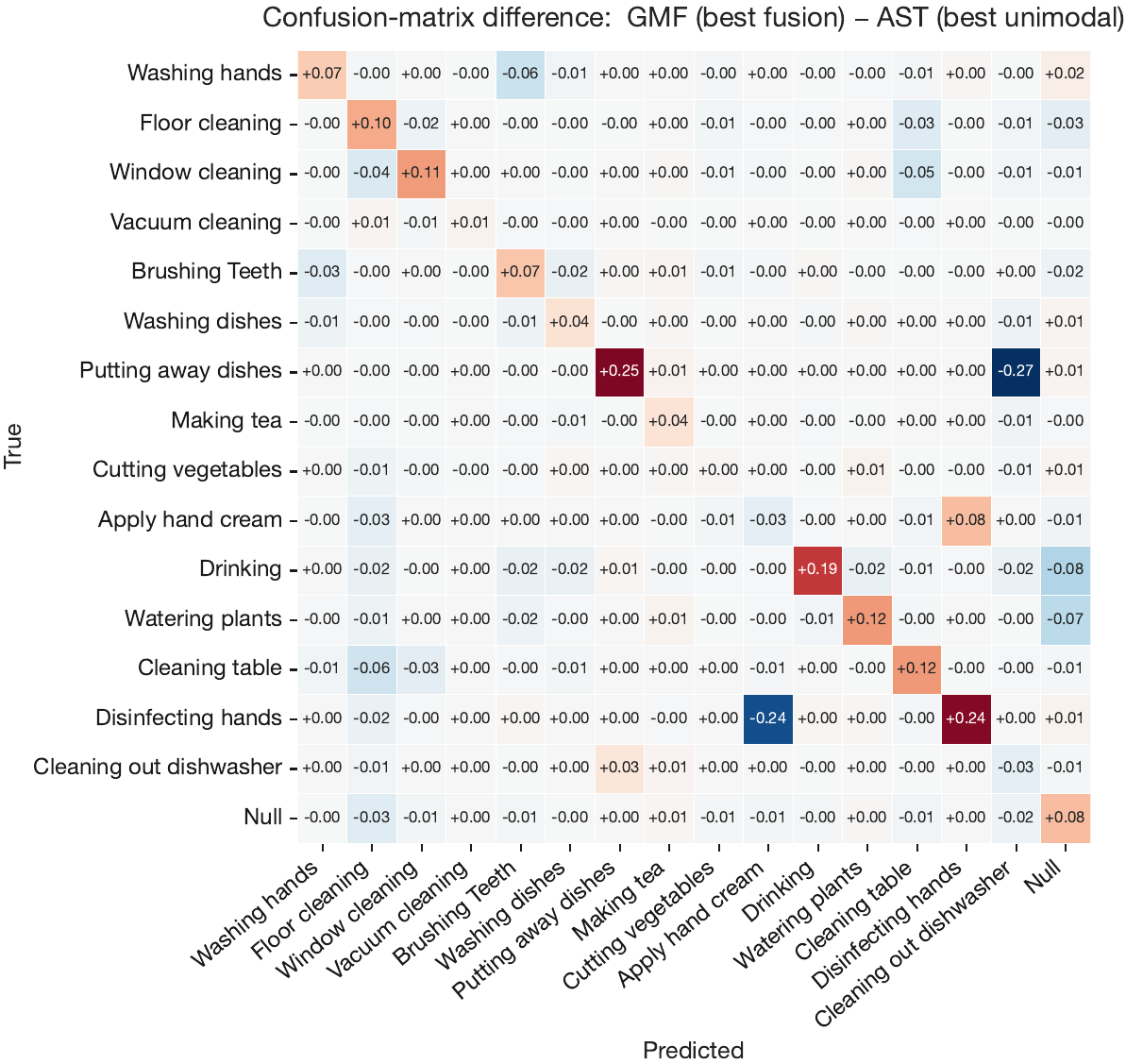}
  \caption{Per-class confusion-matrix difference, GMF minus AST (best unimodal).
  Note the interpretation: Red, positive values on the diagonal mark classes recognised more reliably under fusion. Blue, negative off-diagonal entries mark confusions that fusion removes. Blue, negative values on the diagonal, or off-diagonal red values mean that the GMF model performed worse than the unimodal AST for the specific confusion matrix entry.}
  \label{fig:cm-diff}
\end{figure}

To analyze on which classes the performance is improved by multimodel models, Figure \ref{fig:cm-diff} shows the per-class
confusion-matrix difference between the best multi-modal model (GMF) and the best unimodal one (AST). Each diagonal entry shows how much more often GMF classifies that activity correctly, and the blue, negative off-diagonal entries are the confusions GMF removes in comparison. The improvement is concentrated in activities that are quiet or acoustically ambiguous but involve distinctive hand motion. Notably, \textit{Putting away dishes} is recognized correctly +25pp more often, disinfecting hands +24pp more, and drinking +19pp more. These are exactly the classes where audio alone is weak, as there is little characteristic sound. The overall confounding pair of \textit{Apply hand cream} and \textit{Disinfecting hands} is handled better by GMF as well, although notably, while the confusion of \textit{Disinfecting hands} with \textit{Apply hand cream} is reduced by -24pp, vice-versa, it is increased by +8pp.

We additionally report per-modality absolute confusion matrices, together with those of GMF and the GMF-minus-TinyHAR (multimodal-minus-IMU-only) difference, in Appendix \ref{assec:conf}.

\begin{figure}[h!]
    \centering
    \includegraphics[width=1\linewidth]{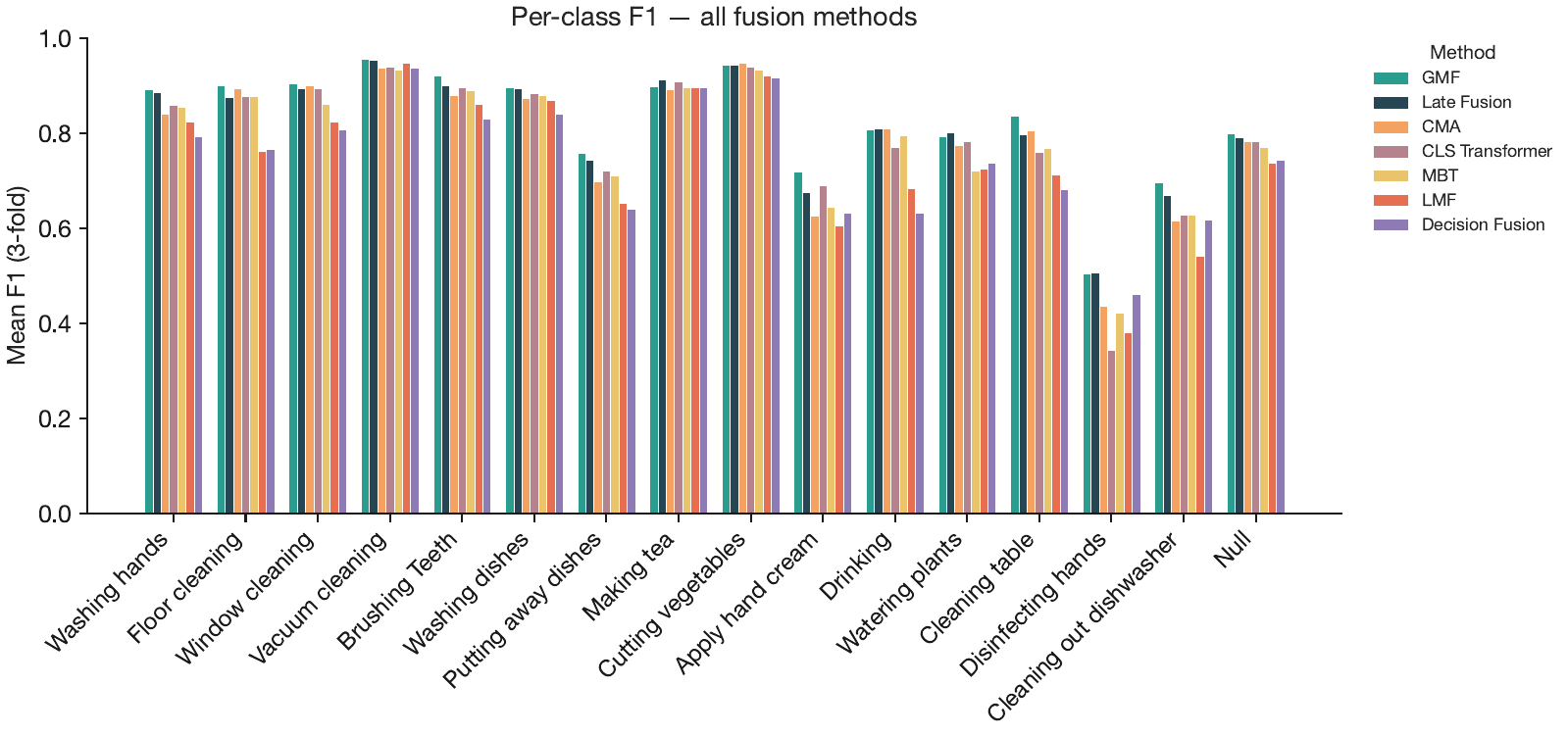}
    \caption{Per-class macro F1 scores for each fusion strategy. Results presented are averaged over the three-fold CV. Models are sorted in descending order by their global performance, from left to right.}
    \label{fig:per-class_fusion}
\end{figure}

In Fig. \ref{fig:per-class_fusion}, we display the per-class F1-scores of all fusion strategies in a bar chart. For each class on the x-axis, the fusion strategies are ordered by their previously calculated macro performance. With minor exceptions, the global ordering of fusion strategy performances also applies to the per-class performance, where GMF and Late Fusion by concatenation perform best. Disinfecting hands is consistently the most challenging class for all fusion strategies.

\subsection{Generalisation to unseen participants}

\subsubsection{Leave-one-participant-out}
\label{sec:res-lopo}

The 3-fold protocol mixes participants across folds. To approximate the performance on a single, unseen participant and to be comparable to the HARMES baseline \cite{burchardHARMESMultiModalDataset2026a}, we re-evaluate the best model, GMF, under the stricter leave-one-participant-out (LOPO) scheme. GMF reaches 0.819 macro F1 under LOPO, within one point of its 3-fold score of 0.827. The drop is negligible despite LOPO being the more demanding test. GMF under LOPO also exceeds the multi-modal HARMES baseline of 0.760 by 5.9 points, on the exact same evaluation protocol. To the best of our knowledge, our multi-modal classification model with GMF is therefore also the currently highest performing model on the HARMES dataset.

The per-class picture reinforces this. We display the GMF LOPO confusion matrix in Fig. \ref{fig:lopo-cm}. When comparing the LOPO confusion matrix with the 3-fold matrix in Appendix \ref{assec:conf} (Figure \ref{afig:cm_gmf}), LOPO matches or slightly exceeds the 3-fold result on most activities, gaining a little on vacuum cleaning, brushing teeth, and window cleaning. The model is almost a 100\,\% accurate on the activities with a clear acoustic and motion signature: vacuum cleaning (0.97), cutting vegetables (0.94), making tea and brushing teeth (0.91), and washing dishes and window cleaning (0.90). Where LOPO falls slightly behind, the gap is confined to the low-support self-care classes, disinfecting hands and applying hand cream, which have the fewest windows and the least distinctive signatures and are therefore the most sensitive to an unseen participant's personalized motion. These differences are small, and the bulk of the activities are recognized cleanly, confirming that the multi-modal model generalizes well to new users rather than overfitting to the training participants.

\begin{figure}[h!t]
  \centering
  \includegraphics[width=\linewidth]{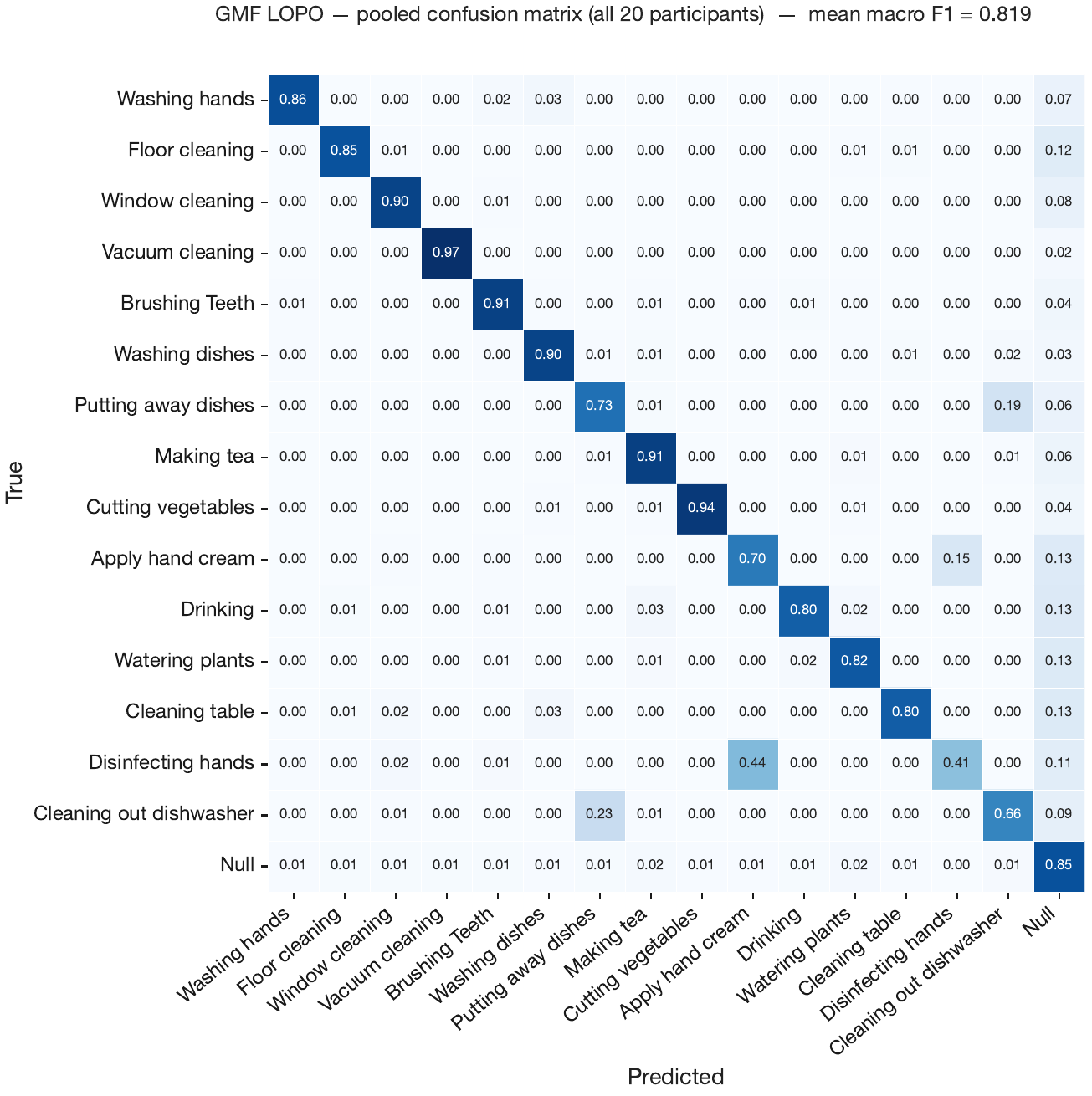}
  \caption{Pooled confusion matrix for GMF under leave-one-participant-out
  evaluation, aggregated over all 20 held-out participants.}
  \label{fig:lopo-cm}
\end{figure}

\subsubsection{Per-participant results and left-handed participants}
\label{sec:res-per-part}
\begin{figure}
    \centering
    \includegraphics[width=1\linewidth]{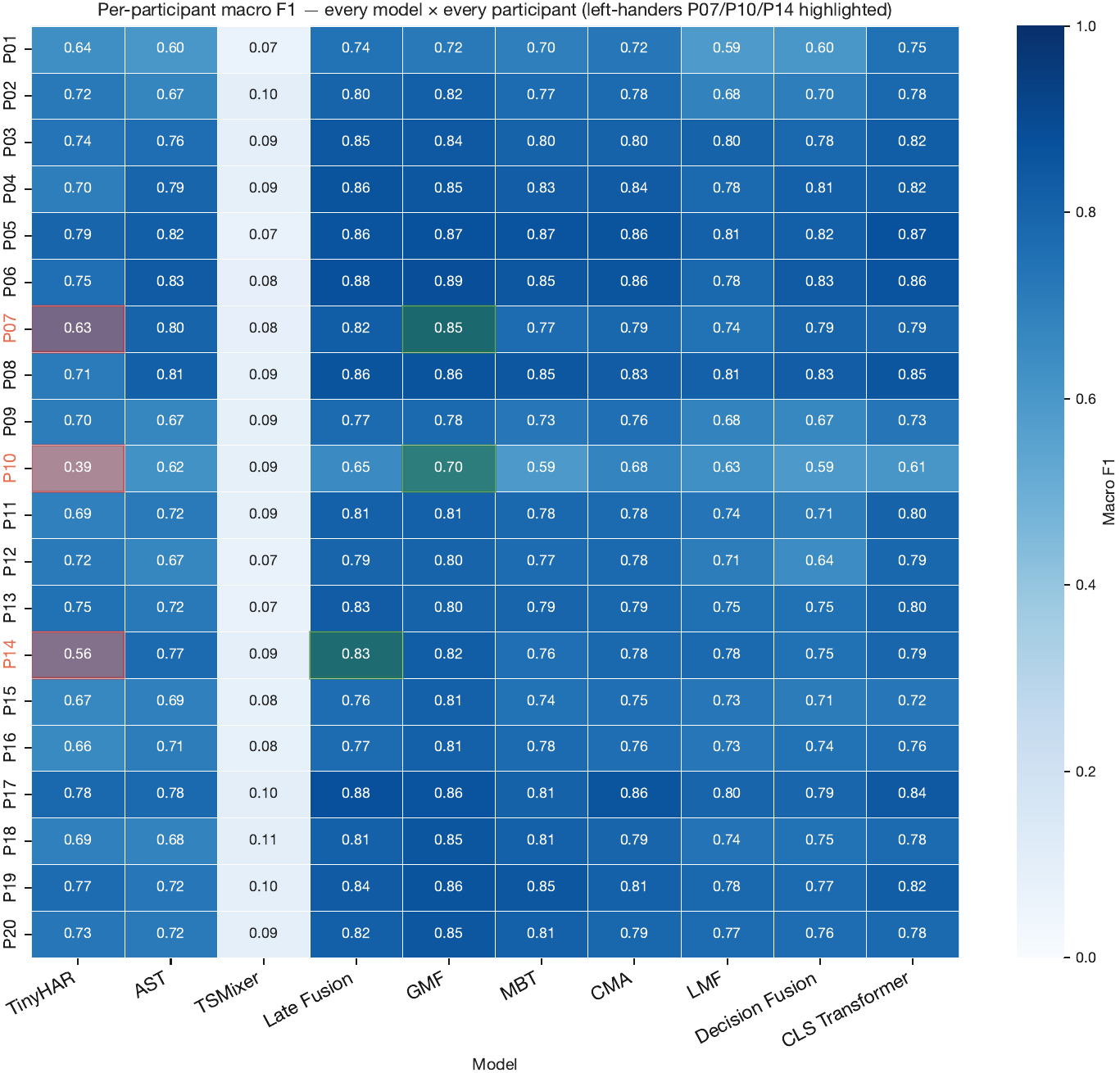}
    \caption{Heatmap table showing per-participant macro F1 scores for each participant and model (3-fold CV). Left-handed participants are marked in red font. The three leftmost models are unimodal: TinyHAR (IMU), AST (Audio), TSMixer (humidity). The worst results on each left-handed participant are marked in red, and the best results on them are marked in green.}
    \label{fig:participant_model_hm}
\end{figure}

The results are mostly stable across participants, but deviations exist. Fig.~\ref{fig:participant_model_hm} shows a heatmap of the macro F1 scores per model and per participant. The largest variations exist in the unimodal IMU-model TinyHAR, with left-handed participants scoring significantly lower. No model strictly beats all other models over all participants. Rather, the results are close, and GMF performs best for most participants, closely followed by Late fusion by concatenation.

\begin{figure}
    \centering
    \includegraphics[width=1\linewidth]{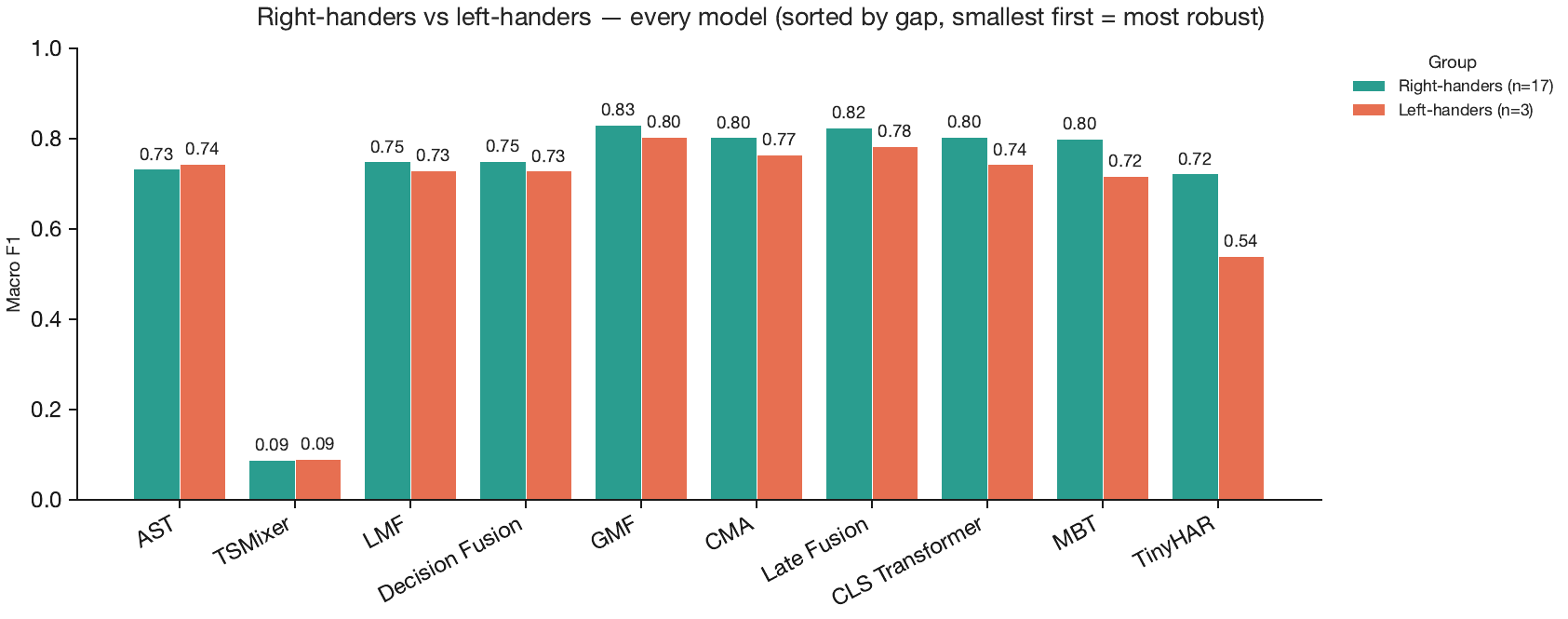}
    \caption{3-Fold CV macro F1 performance per model, split by dominant hand into left-handers and right-handers. The plot shows both unimodal (AST: audio, TSMixer: humidity, TinyHAR: IMU) and multi-modal methods (all others), sorted by performance gap between groups of left-handers and right-handers, descending from left to right.}
    \label{fig:lh-rh-per-model}
\end{figure}

Three participants (P07, P10, P14) are left-handed, which negatively affects the IMU modality
because the dominant-hand motion appears on the opposite wrist from the right-handed
majority. Audio and humidity are largely unaffected by handedness, so any modality-induced robustness gap likely
comes from the IMU. We show the gap between the groups of left-handers and right-handers in Fig. \ref{fig:lh-rh-per-model}, for each fusion strategy and for the unimodal models. The IMU-based unimodal TinyHAR by far shows the largest gap of 0.18 (0.54 vs 0.72).

Humidity is excluded from this comparison because it carries almost no discriminative
signal for any participant.

Audio behaves in the opposite way. AST is essentially handedness-invariant and in fact
performs marginally better on the left-handers (0.74 versus 0.73), since a microphone
picks up the same activity regardless of which hand performs it. Fusion inherits this
robustness and recovers what the IMU alone loses. For every left-hander, the most
informative model is a fusion method.

\section{Discussion}
\label{sec:discussion}
\paragraph{Simple fusion outperforms complex fusion}
From the performance results, we infer that two of the simplest fusion methods score highest. GMF (0.827) and Late
Fusion (0.817) outrank the attention- and transformer-based methods (CMA, CLS
Transformer, MBT, all near 0.79), and the tensor and decision-level methods (LMF,
Decision Fusion) trail at 0.75. This runs against the intuition that richer
cross-modal interaction should help. We attribute it to the setting rather than to a
flaw in those methods. With only three modalities, one of which (humidity) is of limited use in 10s windows, and with a frozen, already-strong audio encoder, there is little structure left for elaborate attention to exploit, and the extra parameters are harder to fit on a dataset of 20 participants, despite the comparatively large size of 61 labeled hours. Especially for HAR, where datasets are often not that large, it seems that these simpler methods already deliver the best performance. 

\paragraph{Fusion improves performance by resolving specific confusions, not by a uniform lift}
The per-class analysis (Section \ref{sec:res-perclass}) shows the performance gain is concentrated
in activities that are ambiguous in a unimodal IMU or audio setting, but have an additional distinctive acoustic or motion pattern. This applies, e.g., to putting away dishes, disinfecting hands, and drinking. These improve because fusion untangles confusion pairs the audio or IMU model could not separate on its own, for example, putting away dishes versus cleaning out the dishwasher, or disinfecting hands versus applying hand cream. Activities with a loud, unmistakable acoustic signature (vacuum
cleaning, cutting vegetables, making tea) are already solved by the acoustic model alone, and their performance thus does not benefit from other modalities. The gain of the fusion models over the best unimodal model (AST) is moderate in total, and fusion adds value precisely on the subset of activities where a microphone is blind. Overall, the results of this work strengthen pre-existing findings that showed the benefits of multi-modal HAR.

\paragraph{Humidity contribution is minimal}
In our 10\,s-window model setting, across the fusion methods, removing the humidity stream changes macro F1 by a fraction
of a percentage point, and TSMixer (0.088) performs near the chance level. Only watering plants
and cleaning out the dishwasher, where moisture genuinely changes, benefit at the class
level. The practical implication is that a two-sensor IMU-plus-audio system captures
essentially all of the achievable performance on this dataset for 10\,s windows, and the humidity sensor
can be dropped without cost. We report this as a negative result worth recording rather than a limitation of the fusion methods. As acknowledged by Burchard et al. in the dataset publication \cite{burchardHARMESMultiModalDataset2026a}, the humidity sensor reacts to the environment changes slowly and should thus be treated differently from the accurate high-frequency measurements of IMU and microphone.

\paragraph{The models generalise to unseen participants}
Under the stricter leave-one-participant-out protocol, where every test participant is
entirely unseen and evaluated alone, GMF reaches 0.819, within one point of its 3-fold
score and 5.9 points above the already multi-modal HARMES baseline of 0.760 on the same protocol.
The per-class behaviour is stable between the two protocols, with the small residual
drops confined to the low-support classes \textit{Apply hand cream} and \textit{disinfecing hands}. This indicates the multi-modal model learns activity signatures that transfer across individuals rather than fitting the particular participants it was trained on. While the literature sees LOPO-validation as the gold standard, the small performance difference between 3-fold CV and LOPO for GMF hints at the validity of the tradeoff between cross-validation meaningfulness and training time/energy consumption we chose.

\paragraph{Fusion improves robustness to handedness}
The IMU-only unimodal model (TinyHAR) degrades sharply for the three left-handed participants, since the
dominant-hand motion appears on the opposite wrist, giving it the largest
right-hand-versus-left-hand gap of any model (0.72 versus 0.54). Audio is
handedness-invariant and is unaffected. Because fusion combines the two, the multimodal
models inherit the audio model's robustness and recover most of what the IMU loses: GMF
narrows the gap to 0.027 and is the best model for the left-handers in nearly every case.
Beyond raw accuracy, this is a fairness argument for multi-modal sensing: a single-modality
IMU system would systematically underperform for a minority of users, and adding audio
removes most of that disparity. Likely, this issue can also be overcome by unimodal IMU models to a certain degree, e.g., by applying sophisticated data augmentation techniques. 

\paragraph{Limitations}
Several caveats bound these conclusions. The dataset has twenty participants and only
three left-handers, so the handedness finding, while consistent, rests on a small sample.
All results come from a single dataset of activities of daily living, and the audio
backbone is frozen, so the ceiling we observe partly reflects a fixed pretrained
representation rather than end-to-end optimization. Due to the design of the encoder networks, the conclusions we draw are specific to this architecture and should be tested against other encoder architectures and datasets in the future. 

The ordering of fusion methods may
also shift with more modalities or larger data, where the more expressive architectures
could have room to pay off. Finally, performance on the rare self-care classes remains modest under cross-participant evaluation, which is the most likely target for future improvement. Our selection of fusion methods is also limited to methods fusing at a late intermediary stage in the model, after passing the data through completely separate encoder heads. Early fusion approaches are excluded from this particular work, but should also be considered by practitioners when developing fusion models. 

\section{Conclusion}
\label{sec:conclusion}
We compared seven fusion strategies for multi-modal activity recognition on the HARMES
dataset under a leakage-free, participant-grouped protocol. Every method improved on the
best unimodal model, and the simplest mechanisms were the strongest: gated multimodal
fusion (GMF) led at 0.827 macro F1, with late concatenation fusion close behind, while the heavier
attention- and tensor-based methods trailed. The multi-modal gains came from resolving confusions between activities that sound alike but display different movement patterns and vice-versa, with audio and IMU proving complementary. The GMF-based model generalized well to unseen participants (F1 of 0.819 under leave-one-participant-out, 5.9 points above the dataset baseline) and was far more robust to handedness than the IMU-only model, suggesting that a lightweight gated fusion of IMU and audio is both the most accurate and the most practical choice on the HARMES dataset.

\label{sec:future}
When considering future applications and enhancements, a natural next step is wearable deployment: quantising and pruning the IMU-plus-audio model for on-device use, with streaming inference for continuous rather than window-level
recognition. On the other hand, while our study already compares seven fusion strategies, more methods, especially from the area of early fusion, could be explored. Our analysis should also be extended to other multi-modal datasets, to strengthen and compare these findings and rule out the influence of a single dataset's specific properties.
For sensing, a longer context or multi-rate fusion would better match humidity's slow dynamics, and adding further sensors or more data would test whether the preference for simple fusion persists as the modality count grows or the dataset size increases.
\begin{credits}

\subsubsection{\ackname} The authors would like to thank the participants of the HARMES dataset and all researchers involved in creating it.

\subsubsection{\discintname}
The authors have no competing interests to declare that are
relevant to the content of this article.

\subsubsection{Contribution Statement.}
AM and RB contributed equally to this work.
AM co-designed the experiments, executed the experiments, collected all results, and co-wrote the manuscript.
RB co-designed the experiments and co-wrote the manuscript.
All authors reviewed the manuscript.

\end{credits}
%
%
%
\bibliographystyle{splncs04}
\bibliography{bibliography_better_bibtex}
\appendix

\section{Fusion Strategy Visualization}
\label{asec:fusion_strategies}
The following diagrams depict the seven fusion strategies we employed for this comparison study. They replace the block ``Fusion Method'' in Fig. \ref{fig:fusion_architecture_overview}.

\begin{figure}
    \centering
    \includegraphics[width=1\linewidth]{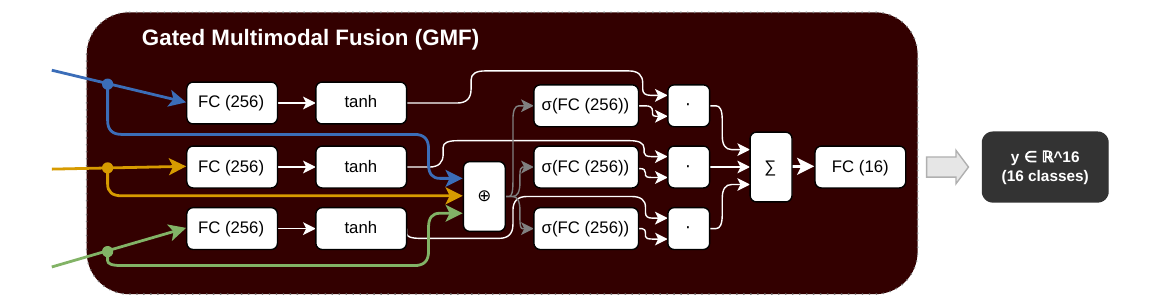}
    \caption{Gated Multi-modal Fusion (GMF)}
    \label{afig:gmf}
\end{figure}

\begin{figure}
    \centering
    \includegraphics[width=1\linewidth]{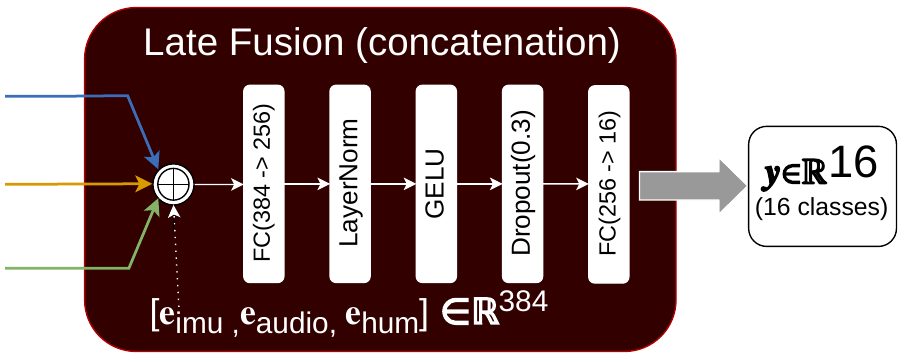}
    \caption{Late Fusion (concatenation)}
    \label{afig:lf}
\end{figure}

\begin{figure}
    \centering
    \includegraphics[width=1\linewidth]{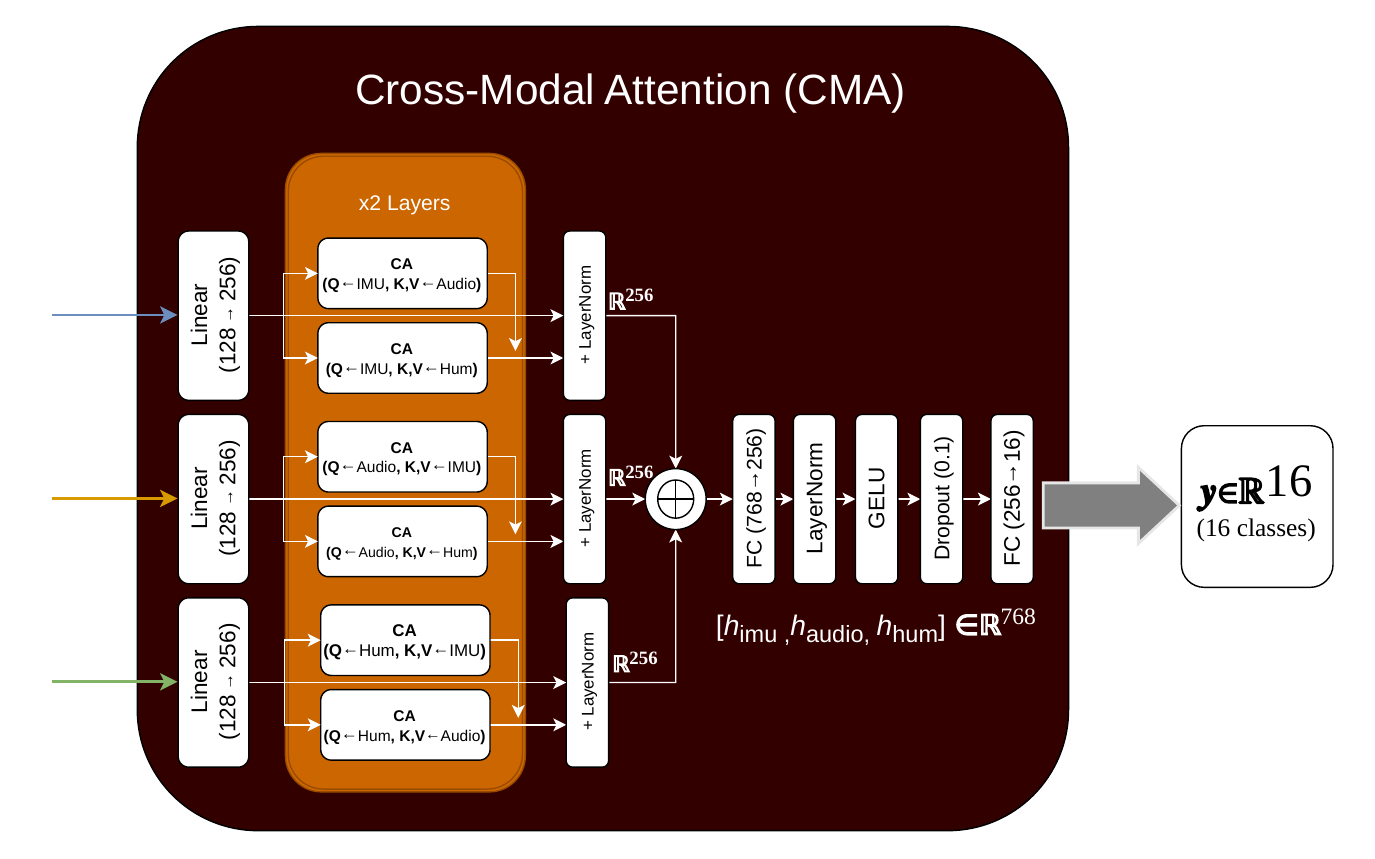}
    \caption{Cross-Modal Attention (CMA)}
    \label{afig:cma}
\end{figure}

\begin{figure}
    \centering
    \includegraphics[width=1\linewidth]{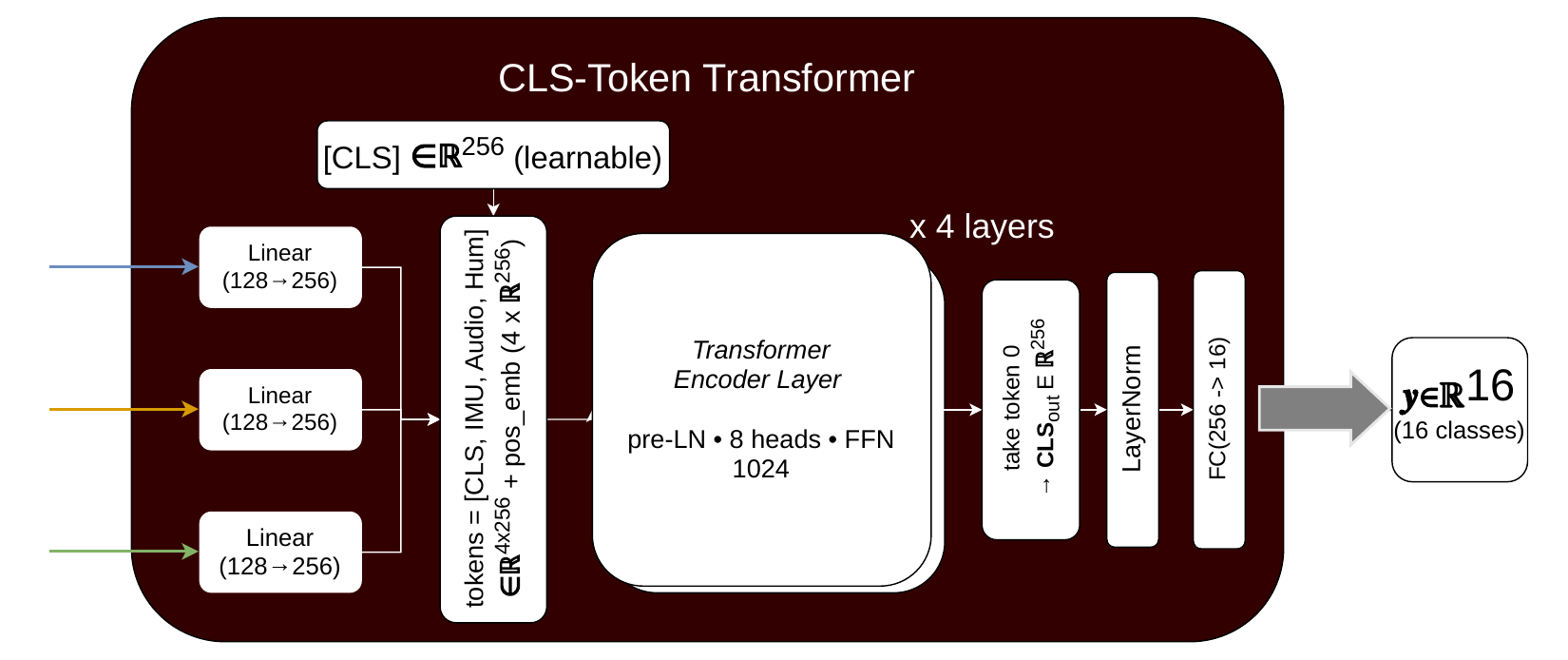}
    \caption{CLS-Token Transformer}
    \label{afig:clstranf}
\end{figure}

\begin{figure}
    \centering
    \includegraphics[width=1\linewidth]{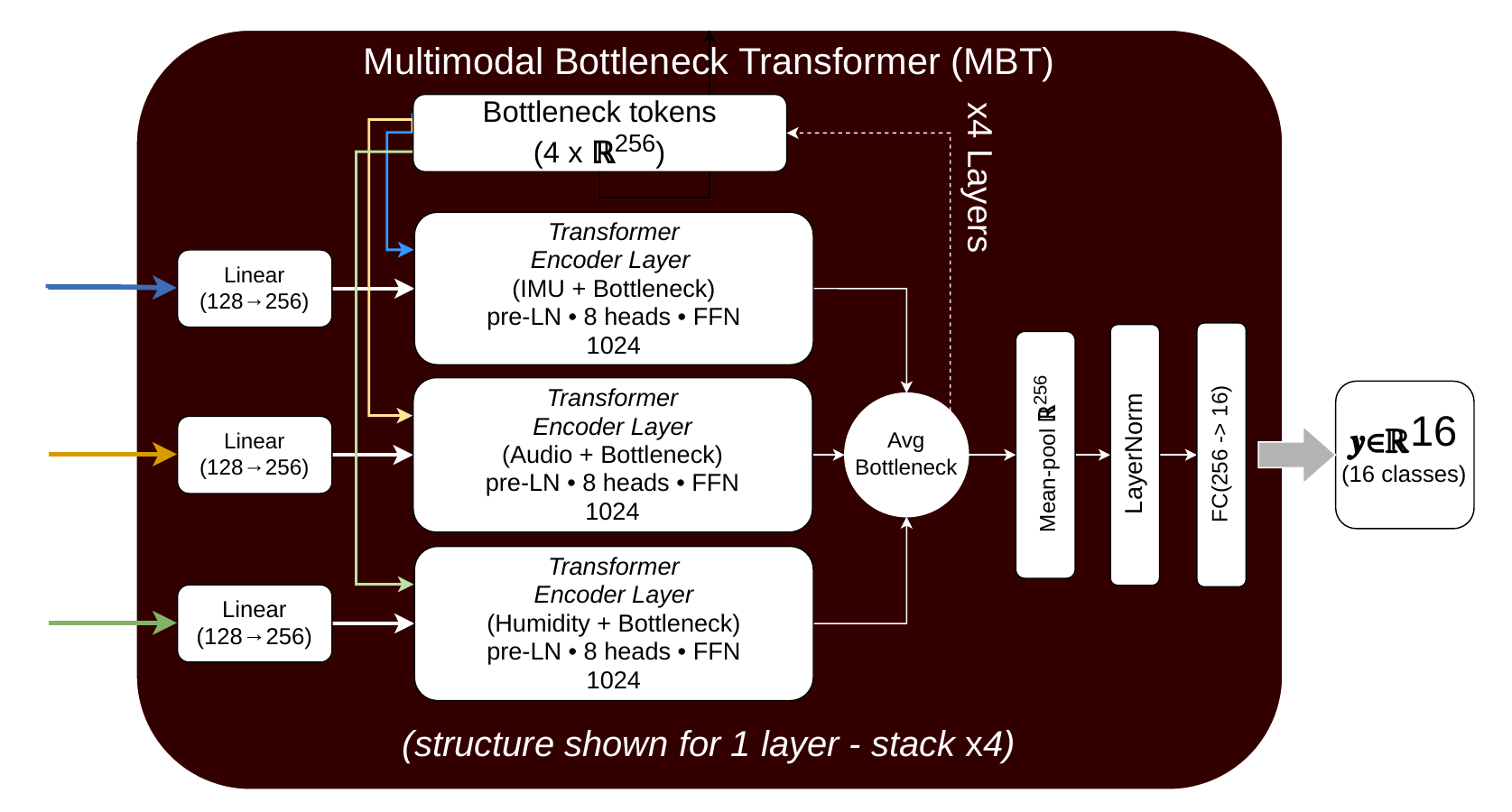}
    \caption{Multi-modal Bottleneck Transformer}
    \label{afig:mbt}
\end{figure}

\begin{figure}
    \centering
    \includegraphics[width=1\linewidth]{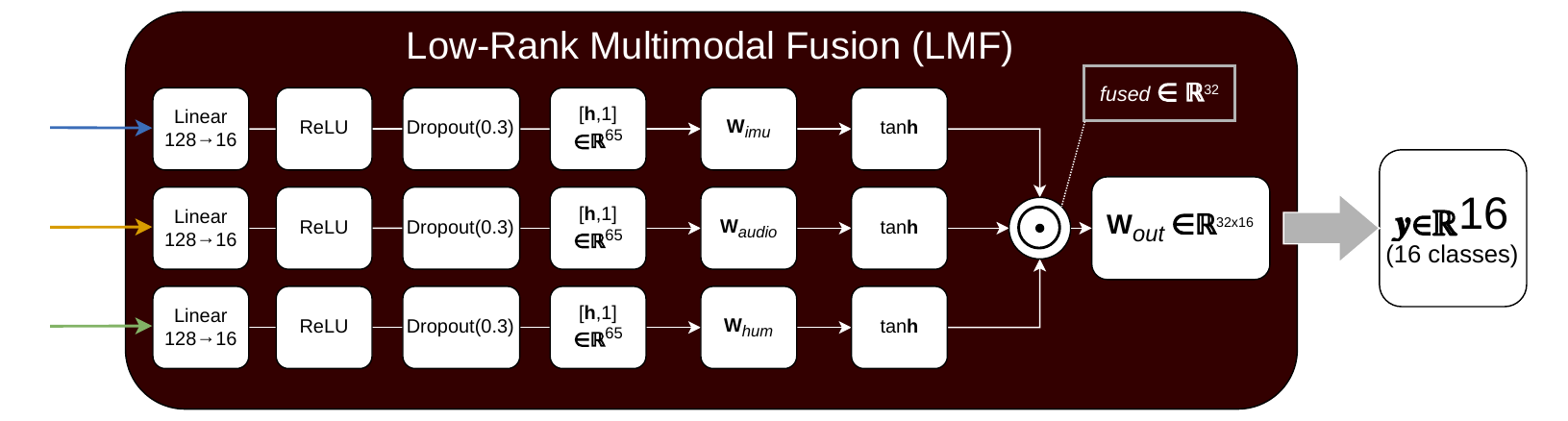}
    \caption{Low-Rank multi-modal Fusion}
    \label{afig:lmf}
\end{figure}

\begin{figure}
    \centering
    \includegraphics[width=1\linewidth]{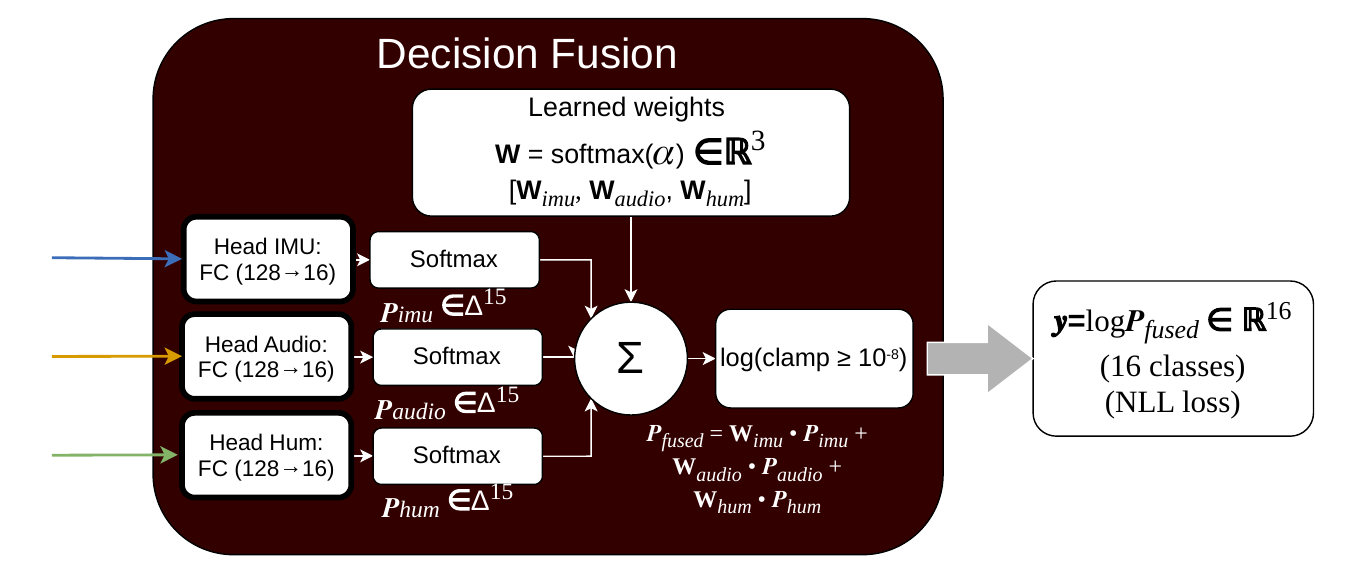}
    \caption{Decision Fusion}
    \label{afig:df}
\end{figure}

\FloatBarrier

\section{Additional Machine Learning Results}
\label{asec:add_ml_results}

\subsection{Confusion Matrices}
\label{assec:conf}
In this section, we show additional confusion matrices. We include one for each unimodal model (AST, TinyHAR, TSMixer), as well as the 3-Fold confusion matrix of the best performing model (GMF), and the confusion difference between TinyHAR and GMF

\begin{figure}
    \centering
    \includegraphics[width=.8\linewidth]{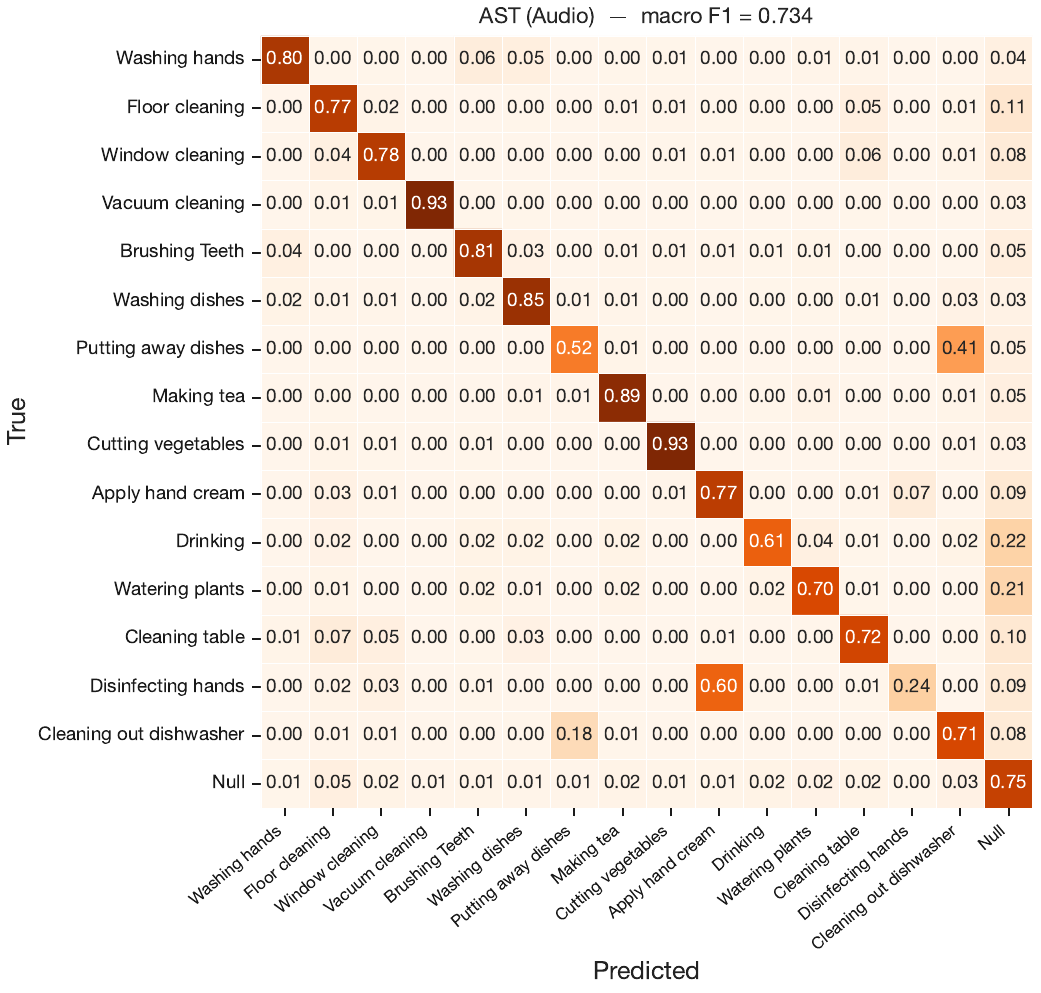}
    \label{afig:cm_ast}
\end{figure}

\begin{figure}
    \centering
    \includegraphics[width=.8\linewidth]{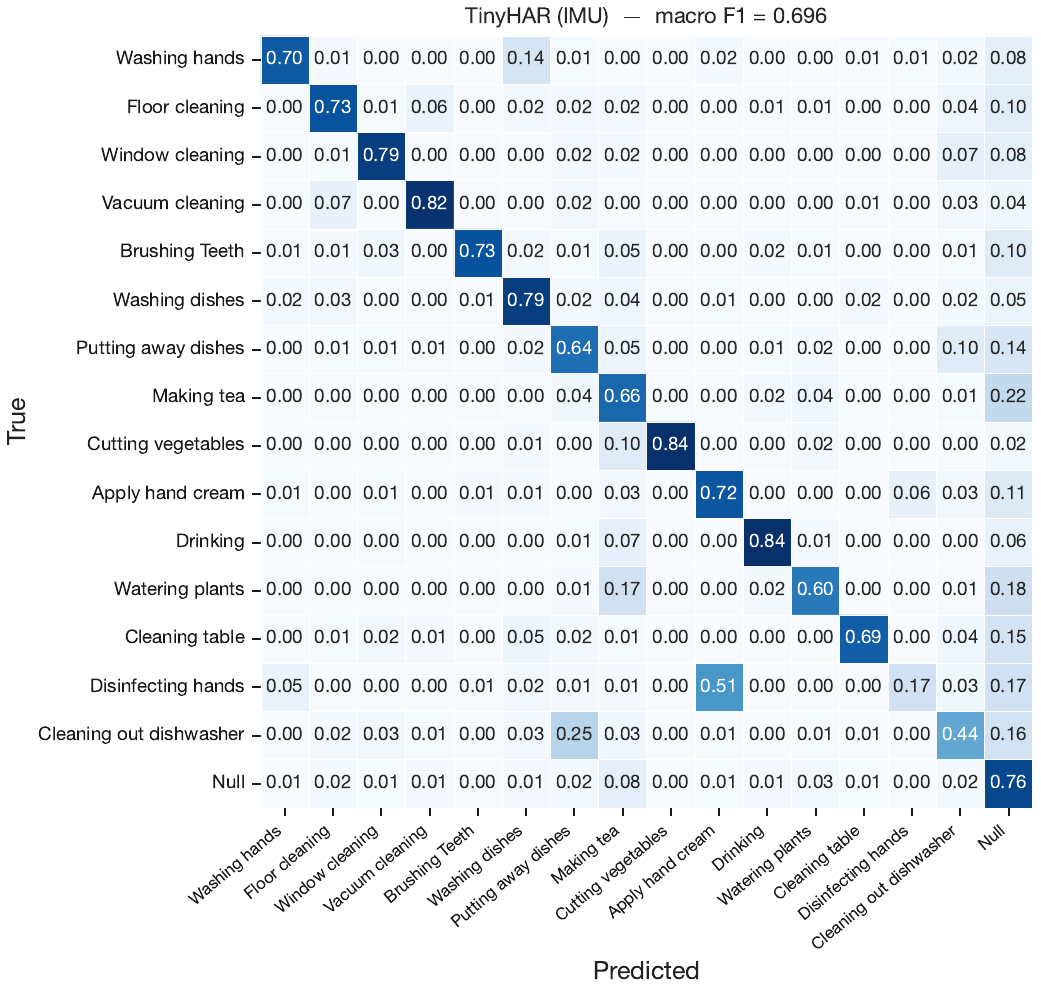}
    \label{afig:cm_tinyhar}
\end{figure}

\begin{figure}
    \centering
    \includegraphics[width=.8\linewidth]{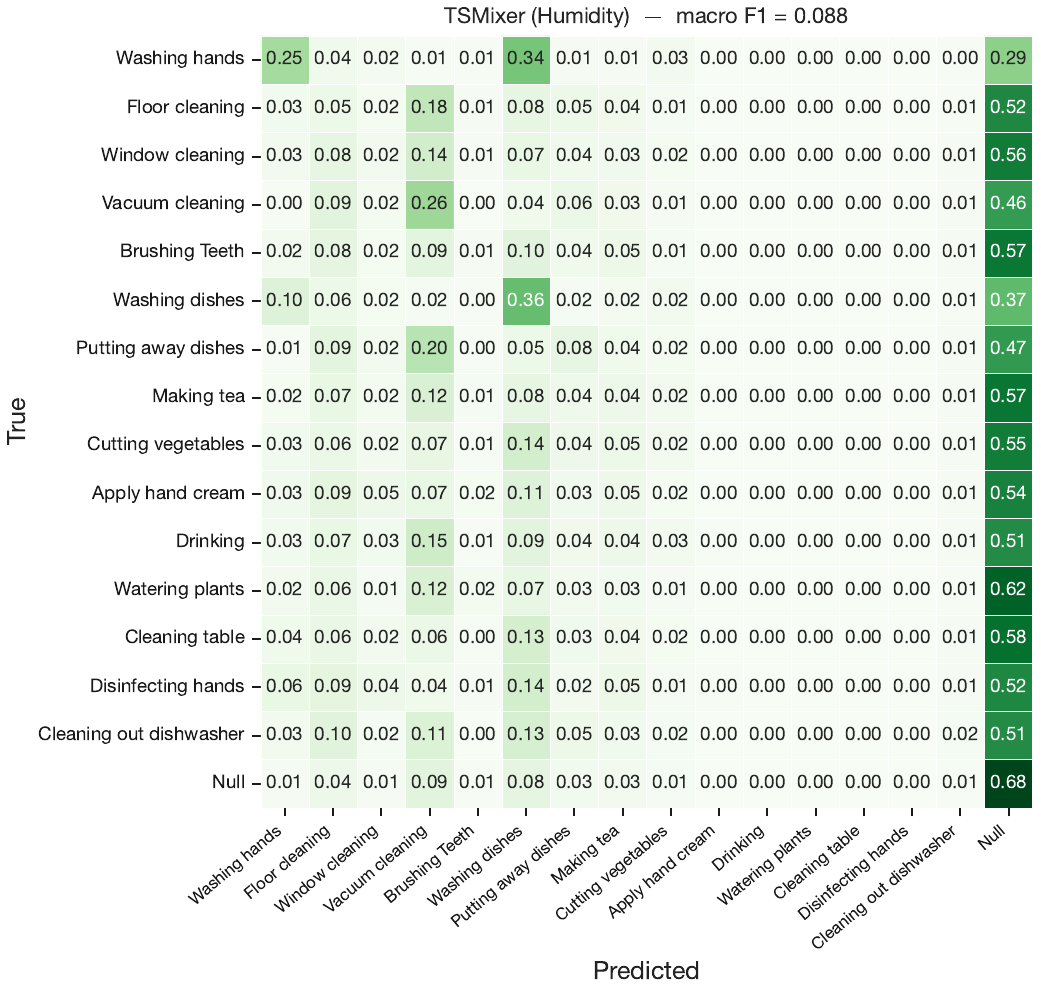}
    \label{afig:cm_tsmixer}
\end{figure}

\begin{figure}
    \centering
    \includegraphics[width=.8\linewidth]{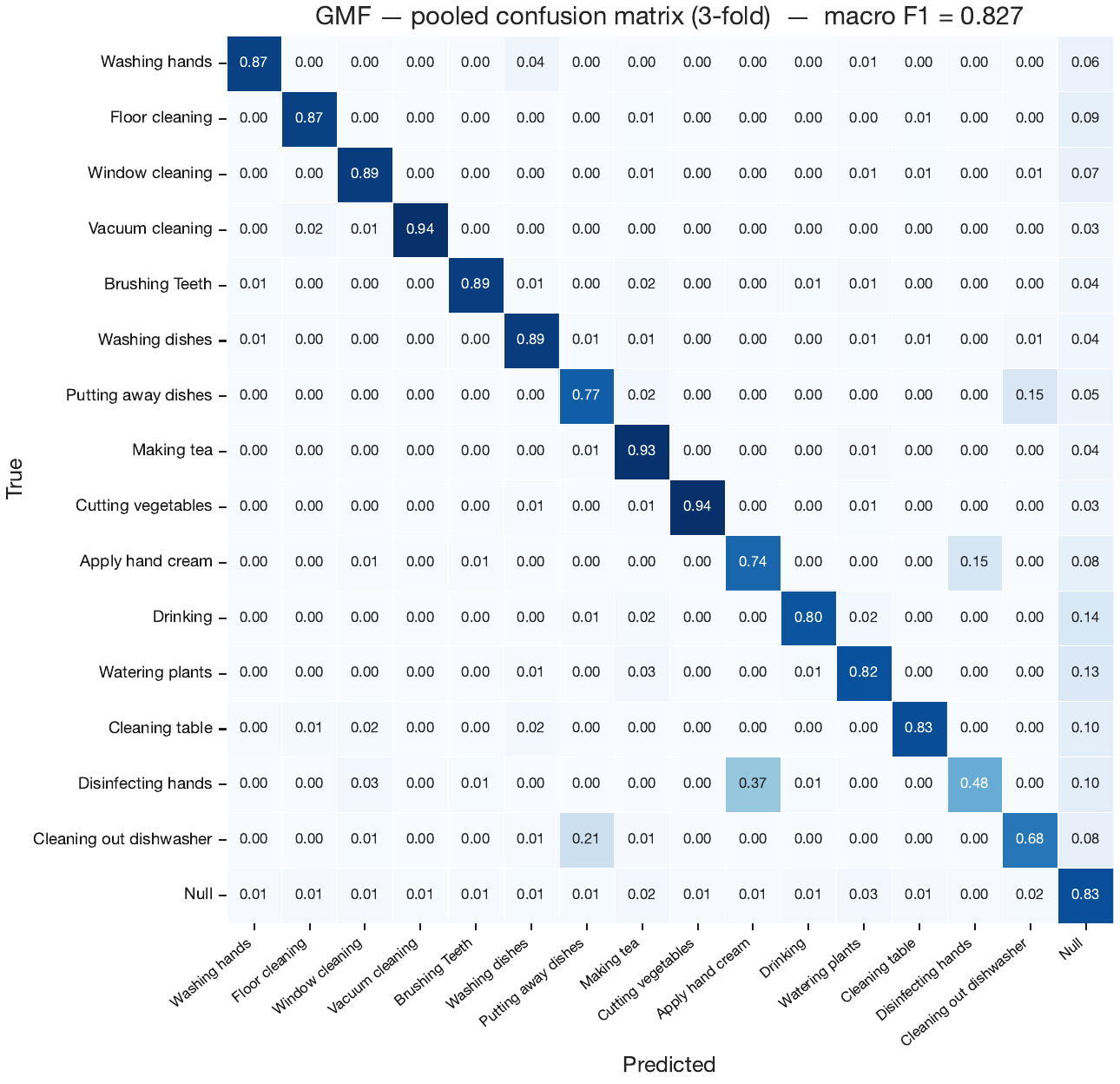}
    \label{afig:cm_gmf}
\end{figure}

\begin{figure}
    \centering
    \includegraphics[width=.8\linewidth]{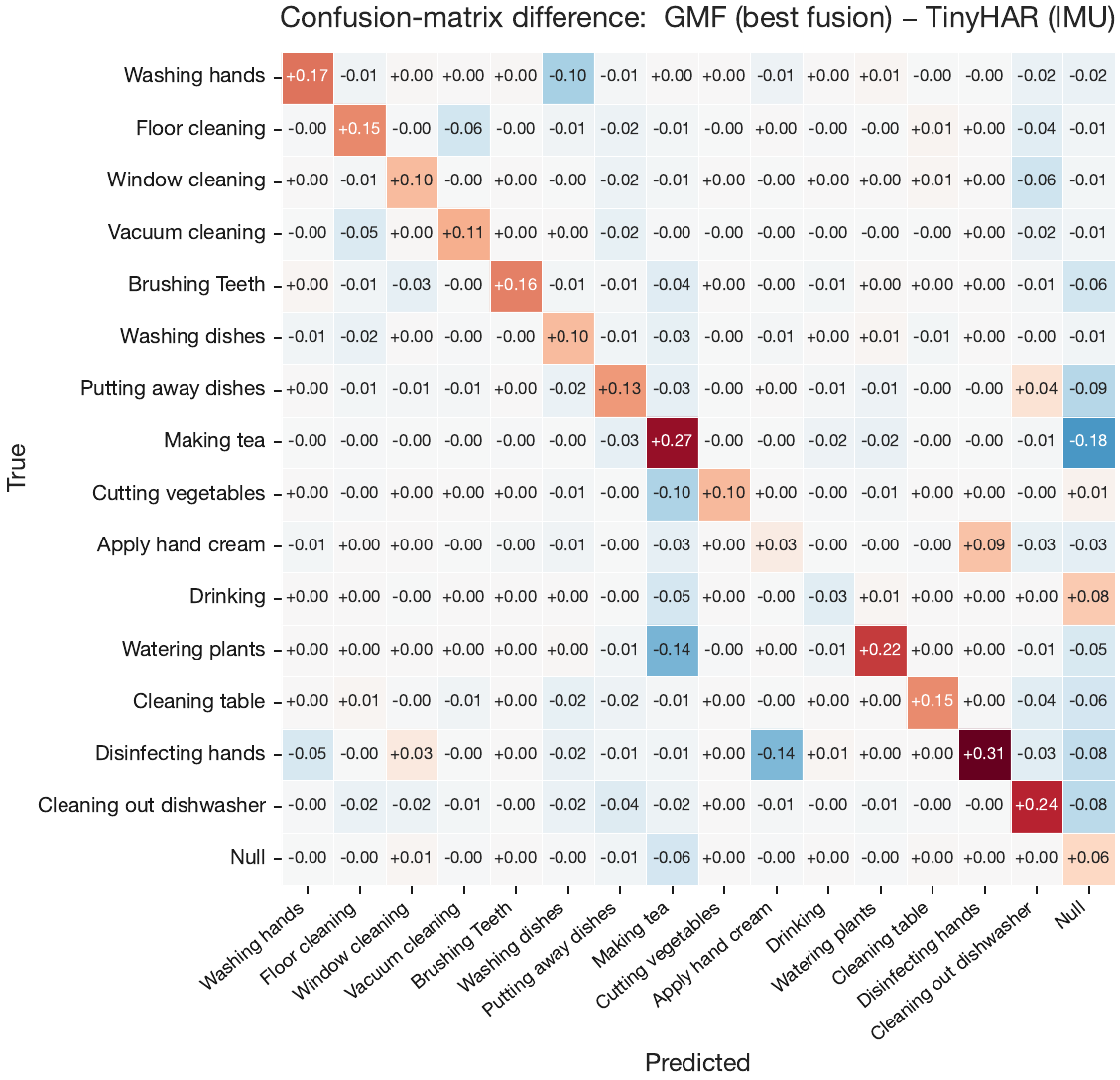}
    \label{afig:cm_diff_gmf_tinyhar}
\end{figure}

\end{document}